\documentclass{article}
\usepackage{ijcai19}
\usepackage{times}
\usepackage{soul}
\usepackage{url}
\usepackage[hidelinks]{hyperref}
\usepackage[utf8]{inputenc}
\usepackage[small]{caption}
\usepackage{graphicx} 
\usepackage{amsmath}
\usepackage{amsthm}
\usepackage{booktabs}
\usepackage[noend,ruled,linesnumbered]{algorithm2e}
\usepackage{algorithmicx}
\usepackage{algpseudocode}
\usepackage{amssymb}
\usepackage[table]{xcolor}
\usepackage{cancel}

\newtheorem{definition}{Definition}
\theoremstyle{definition}
\newtheorem{example}{Example}
\newtheorem{theorem}{Theorem}
\title {A Framework for Parallelizing OWL Classification in Description Logic Reasoners}
\author{Zixi Quan and Volker Haarslev\\
Department of Computer Science and Software Engineering\\
Concordia University, Montreal, Canada\\
\emails \{z\_qua, haarslev\}@encs.concordia.ca
}

\begin{document}

\maketitle

\begin{abstract}
In this paper we report on a black-box approach to parallelize existing description logic (DL) reasoners for the Web Ontology Language (OWL). We focus on OWL ontology classification, which is an important inference service and supported by every major OWL/DL reasoner. We propose a flexible parallel framework which can be applied to existing OWL reasoners in order to speed up their classification process. In order to test its performance, we evaluated our framework by parallelizing major OWL reasoners for concept classification. In comparison to the selected black-box reasoner our results demonstrate that the wall clock time of ontology classification can be improved by one order of magnitude for most real-world ontologies.
\end{abstract}

\section{Introduction}

A major topic of knowledge representation focuses on representing information in a form that computer systems can utilize to solve complex problems. The selected knowledge representation formalism is descriptions logics (DLs) \cite{DLhandbooka}, which is a family of formal knowledge representation languages. It is used to describe and reason about relevant concepts (terminological knowledge - TBox) and individuals (assertional knowledge) of a particular application domain. The widely used Web Ontology Language (OWL) is based on DLs. One of the reasoning components in DL systems is an engine known as classifier which infers entailed subsumption relations from knowledge bases. Research for most DL reasoners is focused on optimizing classification using one single processing core \cite{automated-reasoning-feature-models,OWL-systems,automated-reasoning-scope-and-limits}. 
Considering the ubiquitous availability of multi-processor and multi-core processing units not many OWL reasoners can perform inference services concurrently or in parallel. 

\nocite{DLhandbooka}
\nocite{automated-reasoning-feature-models}
\nocite{OWL-systems,automated-reasoning-scope-and-limits}

In the past various parallel reasoning methods have been proposed: a distributed reasoning architecture to accomplish reasoning through a combination of multiple ontologies interconnected by semantic mappings \cite{distributed-reasoning}; a research methodology for scalable reasoning using multiple computational resources \cite{scalable-parallel-reasoning}; 
a parallel TBox classification approach to build subsumption hierarchies \cite{Aslani-Haarslev-2010b}; an optimized consequence-based procedure using multiple cores/processors for classification of ontologies expressed in the tractable $\mathcal{EL}$ fragment of OWL \cite{ELK2011}; Meissner \shortcite{parallel-ALC-DL} applied some computation rules in a simple parallel reasoning system; a parallel DL reasoner for $\mathcal{ALC}$ \cite{Wu-Haarslev-2012a,Wu-Haarslev-2013a}; merge-based parallel OWL classification \cite{Wu-Haarslev-2013b}; a rule-based distributed reasoning framework that can support any given rule set \cite{distributed-OWL-reasoning};  a framework to formalize the decision problems on parallel correctness and transfer of parallel correctness, providing semantical characterizations, and obtaining tight complexity bounds \cite{reasoning-on-parallel-system}; a parallel architecture for OWL classification using plug-in reasoners \cite{Quan-Haarslev-2018}.

In this paper, we propose a general parallel reasoning framework which can be used to parallelize the classification process of OWL reasoners. Specifically, we mainly focus on three differences and novelties to speed up the OWL classification process: (i) The use of parallel processing \cite{parallel-processors}, with an increasing number of threads, in combination with an atomic shared-memory data structure that is shared among a pool of processors performing precomputation and classification in parallel. Compared to Aslani and Haarslev \shortcite{Aslani-Haarslev-2010b}, where a small set of threads operated on a shared taxonomy via locking, our architecture can update subsumption relations lock-free in a globally shared taxonomy. In comparison to Wu and Haarslev \shortcite{Wu-Haarslev-2013b} our architecture avoids a multitude of subsumption tests due to shared data. (ii) The adoption of \textit{work-stealing} techniques \cite{work-stealing,scalable-work-stealing,work-stealing-localized} to manage adaptive and automatic load balancing for ontologies with varying degrees of reasoning complexity. Compared to Quan and Haarslev \shortcite{Quan-Haarslev-2018}, less memory and computation is required by avoiding overlaps among partitions, reducing the number of subsumption tests, and applying different parallelization techniques such as full-scale work stealing. (iii) The parallel reuse of major OWL reasoners as black-box subsumption testers. Compared to ELK \cite{ELK2011}, our approach has a better performance when many threads are used and is not restricted to a small subset of OWL.

\nocite{parallel-processors}
\nocite{Aslani-Haarslev-2010b}
\nocite{work-stealing}
\nocite{Wu-Haarslev-2013b}
\nocite{work-stealing}
\nocite{Quan-Haarslev-2018}
\nocite{ELK2011}

These advances allowed us to demonstrate the performance our framework against the selected black-box reasoner by classifying a great variety of ontologies. However, since the efficiency of the subsumption tests is constrained by the black-box reasoner and due to the limitation of our current experimental environment (a total of 60 hyper-threading cores supporting up to 120 threads), our results outperform the black-box reasoner when the size of ontologies is less than 10,000 concepts in most cases. We are expecting much better results if other black-box reasoners can be used (by overcoming compatibility problems between different programming languages) and more parallel resources are available for ontologies of bigger sizes.

\section{Preliminaries}

Terminological axioms include general concept inclusion axioms (GCIs), such as the form $C\sqsubseteq D$ where $C,D$ are concept expressions. A TBox $\mathcal{T}$ consists of a finite set of terminological axioms. $\mathcal{T}$ is satisfiable if there exists an interpretation $\mathcal{I}$ that satisfies all the axioms in $\mathcal{T}$, i.e., $C^{\mathcal{I}} \subseteq D^{\mathcal{I}}$ must hold for every axiom $C \sqsubseteq D$, and then $D$ is called a subsumer of $C$ and $C$ a subsumee of $D$. Such an interpretation $\mathcal{I}$ is called a model of $\mathcal{T}$.  An equivalence axiom of the form $C \equiv D$ is an abbreviation for the GCIs $C \sqsubseteq D$ and $D \sqsubseteq C$ and a disjointness axiom $C\sqcap D \sqsubseteq \bot$ is equivalent to $C \sqsubseteq \neg D$. The classification of a TBox results in a subsumption hierarchy (or taxonomy) of all named concepts, with $\top$ ($\bot$) as the root (bottom). 

In a concurrent system, processes can access a shared data structure at the same time. In order to ensure data consistency and avoid conflicts among multiple processes, {\em atomic operations} from the Java concurrency (multithreading) package is not only lock-free by requiring partial threads for constant progress but also wait-free for updating information \cite{Java-Atomic,Java-lock}. Therefore, using an atomic shared-memory structure ensures that such a concurrent approach is a non-blocking algorithm, which can process and schedule threads simultaneously. 

\SetAlgoNoLine
\IncMargin{1em}
\begin{algorithm}[t]
\SetKwInOut{Input}{input}
\SetKwInOut{Output}{output}
\BlankLine 
\Indm
     \Input {Ontology $\mathcal{O}$, Black-box Reasoner $R$}
\Indp
\textsc{createHalf-MatrixStructure}\\
$T \leftarrow \textsc{createThreadPool}$\\
\While {\textsc{getAllAxioms}}
     {  $A \leftarrow \textsc{AxiomDivision}$ \\
       \For{each axiom $A_{i} \in A$}
      {
      \If{$\textsc{scheduleWork}(T)$}
      {$\textsc{preComputing}(A_{i})$}
 }}
\While{$\textsc{getRemainingPossibleSet}$} 
      {$G \leftarrow \textsc{groupDivision}$ \\
       \For{each group $G_{i} \in G$}
       {
       \If{$\textsc{scheduleWork}(T)$}
       {$\textsc{classificationSubTest}(G_i, T)$}
}
}
$\textsc{computeOntologyTaxonomy}$\\
\Return \\
\BlankLine
\BlankLine
\textbf{procedure} \textsc{createHalf-MatrixStructure}\\
$\;\;\; N_\mathcal{O} \leftarrow \textsc{getAllSatConcepts}$\\
$\;\;$ \For {each concept $C_{i} \in N_\mathcal{O}$}
{$\;\;$\textsc{create} $\mathcal{A}_{\mathcal{C}_i}=\langle\mathcal{C}_i,  \mathcal{S}_i, \mathcal{E}_i, \mathcal{D}_i\rangle$ and $\mathcal{P}_{\mathcal{C}_i}$}
$\;\;\; \textsc{defineOrder}(N_\mathcal{O})$\\
\BlankLine
\BlankLine
\textbf{procedure} \textsc{defineOrder}($N_\mathcal{O}$)\\
$\;\;\; \Return \;\mathcal{C}_{a}\gtrdot \mathcal{C}_{b}\gtrdot ... \gtrdot\mathcal{C}_{c}\gtrdot\mathcal{C}_{d}... \gtrdot\mathcal{C}_{i}\gtrdot\mathcal{C}_{j}$\\  
\BlankLine
\BlankLine
\textbf{procedure} \textsc{scheduleWork}($T$)\\
$\;\;\; T_i\leftarrow\textsc{getAvailableThread}(T)$\\
$\;\;\;  \textsc{startBlack-boxReasoner}(T_i)$\\ 
$\;\;\;  \Return \;T_i$
\BlankLine 
\caption{\textsc{parallelClassification}}
\label{alg1}
\end{algorithm}

\section{Parallel Reasoning}
The goal of our method is to classify and construct the whole taxonomy and balance the allocation of resources and memory simultaneously in an efficient way. When it comes to parallelization, there are two important factors that affect the classification performance: concurrency and locking (waiting time). In order to balance these two problems with the potential occurrence of big-size ontologies and nonuniformity of subsumption tests, we create an atomic half-matrix shared-memory structure to maintain all the updated information with different sets and the parallel classification approach is mainly separated into two phases: precomputing (line 3-7) and classification phase (line 8-12) with black-box reasoners for each thread (line 22-25) in Algorithm \ref{alg1}.

\subsection{Half-Matrix Data Structure}

A shared-memory half-matrix structure $\mathcal{A}$ contains quadruples $\mathcal{A}_{\mathcal{C}_i}$ for each concept $\mathcal{C}_i\in N_\mathcal{O}$ with $N_\mathcal{O} = \{\mathcal{C}_1, \ldots, \mathcal{C}_{n}\}$ containing all satisfiable concepts of an ontology $\mathcal{O}$ (or TBox), where $n$ is the total number of concepts and $\mathcal{P}$ a finite set of potential possible subsumees of all concepts in $N_\mathcal{O}$ (see line 15-19 in Algorithm \ref{alg1}). For all concepts $\mathcal{C}_i \in N_\mathcal{O}$, we use $\gtrdot$ to indicate an arbitrary but fixed order between every pair of concepts (line 20-21). For the pair $\langle\mathcal{C}_i, \mathcal{C}_{j}\rangle \in N_\mathcal{O}$, \textit{if} $\mathcal{C}_i \gtrdot \mathcal{C}_{j}$, \textit{then} all the operations related to $\mathcal{A}_{\mathcal{C}_i}$ and $\mathcal{A}_{\mathcal{C}_{j}}$ operate on the three sets $\mathcal{S}_i, \mathcal{E}_i, \mathcal{D}_i$ in $\mathcal{A}_{\mathcal{C}_i}$ with  $\mathcal{A}_{\mathcal{C}_{j}}$ indexing $\mathcal{A}_{\mathcal{C}_i}$ and its related sets. 

For all the satisfiable concepts in $N_\mathcal{O}$, a half-matrix structure represents all possible relations with other concepts inferred or tested by a black-box reasoner, e.g., $\textsc{subs?}(C_{2},C_{1})$ becomes true if $C_{1} \sqsubseteq C_{2}$. 

\begin{definition}[Concept Information Sets]
A quadruple $\mathcal{A}_{\mathcal{C}_i}=\langle\mathcal{C}_i,  \mathcal{S}_i, \mathcal{E}_i, \mathcal{D}_i\rangle$ contains known information for every satisfiable $\mathcal{C}_i \in N_\mathcal{O}$, where $\mathcal{S}_i$ contains $\mathcal{C}_i$'s direct subsumees, $\mathcal{E}_i$ $\mathcal{C}_i$'s equivalent concepts including itself, and $\mathcal{D}_i$ $\mathcal{C}_i$'s disjoint concepts. A set $\mathcal{P}_{\mathcal{C}_i}$ contains all the possible subsumees of $\mathcal{C}_i$.
\end{definition}

\begin{definition}[Remaining Possible Subsumees] A set $\mathcal{R_O}$ is defined as $\mathcal{R_O} = \bigcup_{\mathcal{C}_{i} \in N_{\mathcal{O}}}\{\mathcal{P}_{\mathcal{C}_i}\}$, which reflects all possible sets $\mathcal{P}_{\mathcal{C}_{i}}$ where $\mathcal{P}_{\mathcal{C}_{i}} \neq \emptyset$. 
\end{definition}

\begin{example} 
Assume we have satisfiable concepts $N_\mathcal{O} = \{C_{1}, C_{2}, C_{3}, C_{4}, C_{5}, C_{6}\}$ and their defined order is ${C_2} \gtrdot {C_3} \gtrdot {C_1} \gtrdot {C_4} \gtrdot {C_5} \gtrdot {C_6}$. The relations among these tested concepts are $\mathcal{O} \models \{C_1 \sqsubseteq C_2$, $C_2 \not\sqsubseteq C_1$, $C_6 \not\sqsubseteq C_3$, $C_3 \sqsubseteq \mathcal{C}_6$\}. Accordingly the changes to $\mathcal{P}$ and $\mathcal{A}$ result in the following sets:

\begin{center}
	\begin{tabular}{ll} 
	 $\mathcal{A}_{C_2} \rightarrow \mathcal{S}_2 = \{C_1\}$ & $\mathcal{P}_{C_2}$ = \{$C_3$, \cancel{$C_1$}, $C_4$, $C_5$, $C_6$\} \\
     	 $\mathcal{A}_{C_3}\rightarrow \mathcal{S}_3 =  \emptyset$ & $\mathcal{P}_{C_3}$ = \{$C_1$, $C_4$, $C_5$, \cancel{$C_6$}\}\\          
     	 $\mathcal{A}_{C_1}\rightarrow \mathcal{S}_1 =  \emptyset$ & $\mathcal{P}_{C_1}$ = \{$C_4$, $C_5$, $C_6$\} \\          
      	 $\mathcal{A}_{C_4}\rightarrow \mathcal{S}_4 = \emptyset$ & $\mathcal{P}_{C_4}$ = \{$C_5$, $C_6$\} \\                                 
      	 $\mathcal{A}_{C_5}\rightarrow \mathcal{S}_5 = \emptyset$ & $\mathcal{P}_{C_5}$ = \{$C_6$\}\\
      	 $\mathcal{A}_{C_6}\rightarrow \mathcal{S}_6 = \{C_3\}$ & $\mathcal{P}_{C_6}$ = $\emptyset$\\
	\end{tabular}      
\end{center}
\end{example}

Therefore, we obtain two subsumption testing results for every pair of concepts, which guarantee completeness and require less memory for updating the changes in $\mathcal{P}$ and $\mathcal{A}$, until all concepts $\mathcal{C}_i \in \mathcal{N_O}$ have been tested.

\subsection{Maintaining Sets}
Given $\mathcal{A}_{\mathcal{C}_i}=\langle\mathcal{C}_i,  \mathcal{S}_i, \mathcal{E}_i, \mathcal{D}_i\rangle$ of $\mathcal{C}_i\in N_\mathcal{O}$, for every pair \{$\mathcal{C}_i, \mathcal{C}_j\} \in N_\mathcal{O}$ with $i \neq j$, all the concepts in both $\mathcal{E}_i$ ($\mathcal{D}_i$) and $\mathcal{E}_j$ ($\mathcal{D}_j$) are disjoint.  The related operations for each type of sets $\mathcal{S}_i, \mathcal{E}_i, \mathcal{D}_i$ of $\mathcal{A}_{\mathcal{C}_i}$ are shown in Algorithm \ref{alg2}.

\SetAlgoNoLine
\begin{algorithm}[t]
\BlankLine 
\textbf{procedure} \textsc{updateSubsumee($C_i, C_j$)}\\
$\;\;$ \If {\textsc{subs?}$(C_i, C_j)$} 
{$\mathcal{S}_i = \mathcal{S}_i \cup \{C_j\}$\\
\If{$\mathcal{S}_j \neq \emptyset$}{\textsc{delete} $C\in\mathcal{S}_j$ in ${P}_{C_i}$}
\If {\textsc{subs?}$(C_j, C_i)$}
{$\mathcal{S}_j = \mathcal{S}_j \cup \{C_i\}$\\
\If{$\mathcal{S}_i \neq \emptyset$}{\textsc{delete} $C\in\mathcal{S}_i$ in ${P}_{C_j}$}
\textsc{updateEquivalent}$(C_i, C_j)$}
\textsc{delete} $C_j$ in $\mathcal{P}_{C_i}$\\
}
$\;\;$ \textsc{delete} $C_i$ in $\mathcal{P}_{C_j}$\\   
\BlankLine
\BlankLine
\textbf{procedure} \textsc{updateEquivalent($C_i, C_j$)}\\
$\;\;\; \mathcal{C}_{a}\leftarrow$\textsc{mappingEquivalent}($C_i$)\\
$\;\;\; \mathcal{C}_{b}\leftarrow$\textsc{mappingEquivalent}($C_j$)\\
$\;\;\; \textsc{checkDefinedOrder}(\mathcal{C}_{a}, \mathcal{C}_{b})$\\
$\;\;$ \If {$\mathcal{E}_{b} \mathcal{\setminus}(\mathcal{E}_{b} \cap \mathcal{E}_{a}) \neq \emptyset$} 
 {$\;\;$\textsc{delete} $\mathcal{E}_{b} \mathcal{\setminus} (\mathcal{E}_{b} \cap\mathcal{E}_{a})$ in $\mathcal{P}_{\mathcal{C}_{a}}$}
$\;\;\; \mathcal{P}_{\mathcal{C}_{a}}=\emptyset$,
$\; \mathcal{E}_{a}=\mathcal{E}_{a}\cup\mathcal{E}_{b}, \;\mathcal{E}_{b}=\emptyset$\\
\BlankLine
\BlankLine 
\textbf{procedure} \textsc{updateDisjoint($C_i, C_j$)}\\
$\;\;\; \mathcal{C}_{c}\leftarrow$\textsc{mappingDisjoint}($C_i$)\\
$\;\;\; \mathcal{C}_{d}\leftarrow$\textsc{mappingDisjoint}($C_j$)\\
$\;\;\; \textsc{checkDefinedOrder}(\mathcal{C}_{c}, \mathcal{C}_{d})$\\
$\;\;$ \If {$(\mathcal{D}_{c}\cup\mathcal{D}_{d})\mathcal{\setminus}(\mathcal{D}_{c}\cap\mathcal{D}_{d}) \neq \emptyset$} 
 { $\;\;$ \textsc{delete} \\ 
 $\;\;\; \mathcal{D}_{d}\mathcal{\setminus}(\mathcal{D}_{d}\cap\mathcal{D}_{c})$ in $\mathcal{P}_{C}$ for $C \in {\mathcal{S}_{c}}$,\\ 
 $\;\;\; \mathcal{D}_{c}\mathcal{\setminus}(\mathcal{D}_{d}\cap\mathcal{D}_{c})$ in $\mathcal{P}_{C}$ for $C \in \mathcal{S}_{d}$}
$\;\;$ \textsc{delete} $C_{d}$ in $\mathcal{P}_{C_{c}}$, $C_{c}$ in $\mathcal{P}_{C_{d}}$ \\
$\;\;\; \mathcal{D}_{c}=\mathcal{D}_{c}\cup\{\mathcal{C}_{d}\}, \;\mathcal{D}_{d}=\mathcal{D}_{d}\cup\{\mathcal{C}_{c}\}$\\
\BlankLine 
\caption{\textsc{maintainSets}}
\label{alg2}
\end{algorithm}

\begin{example} 
\label{ex:example-2}
In an ontology $\mathcal{O}$, there are six satisfiable concepts in $N_\mathcal{O} = \{C_1, C_2, C_3, C_4, C_5, C_6\}$ and their defined order is ${C_2} \gtrdot {C_3} \gtrdot {C_1} \gtrdot {C_4} \gtrdot {C_5} \gtrdot {C_6}$. Assume the following relations among these concepts hold:
$\mathcal{O}\models\{C_1\equiv C_5$, $C_3\equiv C_4$, $C_6\sqsubseteq C_2$, $C_3\sqsubseteq C_6$, $C_2\sqcap C_5 \sqsubseteq \bot\}$. 

Since $C_3\sqsubseteq C_6$, $C_6\sqsubseteq C_2$ and $C_3\equiv C_4$, we can infer that $C_3\sqsubseteq C_2$ and $\{C_3, C_4\} \in \mathcal{S}_{6}$. With reference to \textsc{updateSubsumee} in Algorithm \ref{alg2} (line 1-12), the changes to the subsumee sets are $\mathcal{S}_2 = \{C_6\}$, $\mathcal{S}_3=\emptyset$, $\mathcal{S}_1=\emptyset$, $\mathcal{S}_4=\emptyset$, $\mathcal{S}_5=\emptyset$ and $\mathcal{S}_6=\{C_3, C_4\}$. According to \textsc{updateEquivalent}  (line 13-19), $C_1\equiv C_5$, $C_3\equiv C_4$ and $C_2\gtrdot C_3\gtrdot C_1\gtrdot C_4\gtrdot C_5$ are known, so the changes to the equivalent sets are $\mathcal{E}_2 = \{C_2\}$, $\mathcal{E}_3 = \{C_3, C_4\}$,  $\mathcal{E}_1=\{C_1, C_5\}$, $\mathcal{E}_4=\emptyset$, $\mathcal{E}_5=\emptyset$ and $\mathcal{E}_6 = \{C_6\}$. Because of $C_2\sqcap C_5 \sqsubseteq \bot$ and $C_1\equiv C_5$, $C_2\sqcap C_1 \sqsubseteq \bot$ is inferred. Since $C_3$, $C_4$ and $C_6$ are subsumees of $C_2$, therefore, both $C_1$ and $C_5$ are disjoint with $C_3$, $C_4$ and $C_6$. Based on \textsc{updateDisjoint} (line 20-29), the changes to the disjoint sets are $\mathcal{D}_2=\{C_1, C_5\}$, $\mathcal{D}_3=\emptyset$, $\mathcal{D}_1=\{C_2, C_6, C_3, C_4\}$, $D_4=\emptyset$, $D_5=\emptyset$ and $\mathcal{D}_6=\emptyset$. The complete changes are shown in Figure \ref{Example2}, which indicates that all the subsumption relations among the six concepts have been found and there are no more subsumption tests required, which results in $\mathcal{R_O} = \emptyset$.

\begin{figure}[h]
\centering
\includegraphics[width=.49\textwidth]{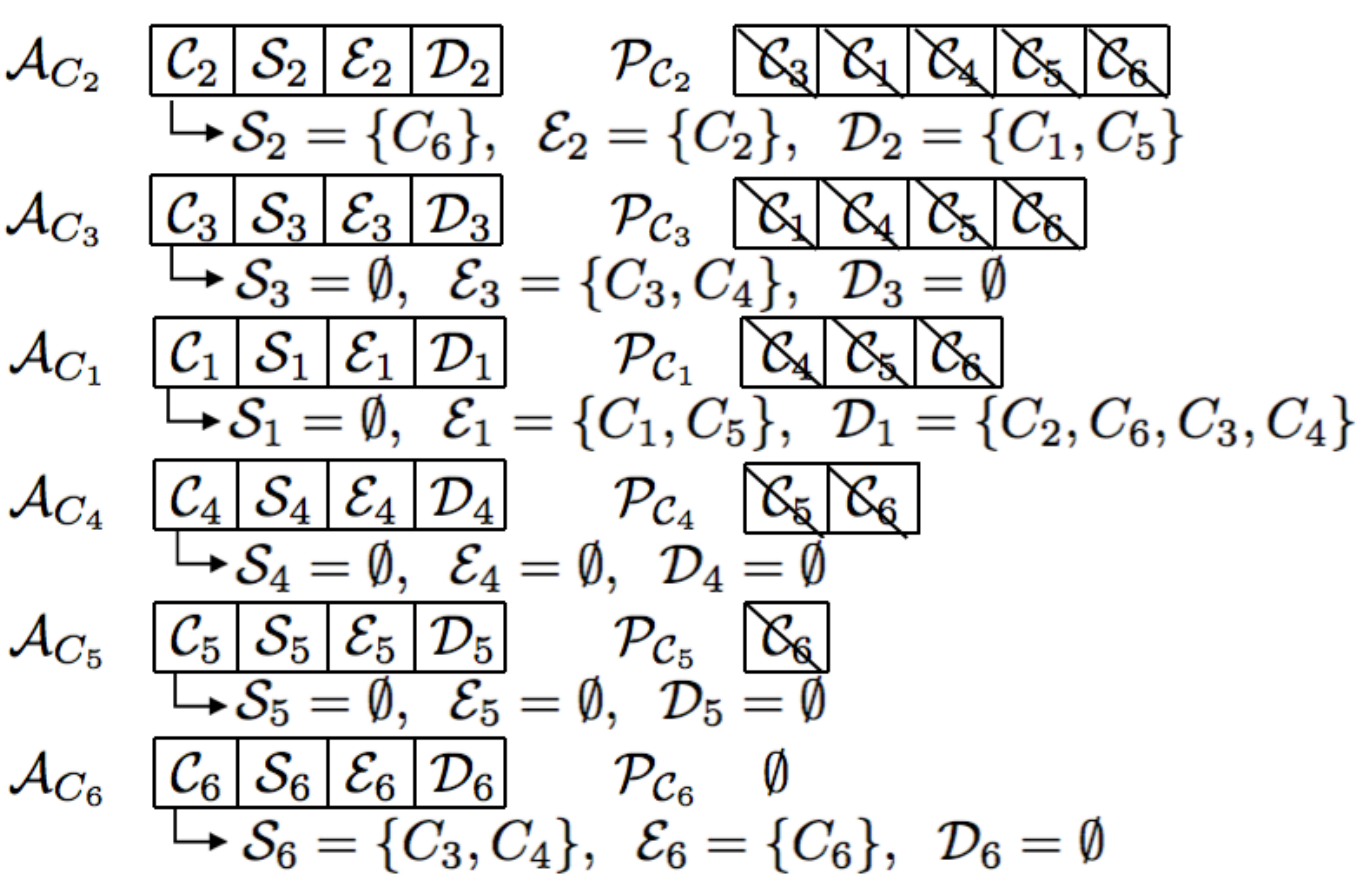}
\caption{Complete changes to $\mathcal{A}$ after applying rules}
\label{Example2}
\end{figure}
\end{example}

\subsection{Precomputing Phase}
In the precomputing part, we use OWL API \cite{OWL-API} to retrieve all declared axioms of an ontology $\mathcal{O}$, and a pool of axioms is created to store these axioms. Whenever a subsumption can be directly derived from an axiom, e.g., $A \sqsubseteq B$, if the converse subsumption is unknown, it is tested using the chosen black-box reasoner, e.g., $\textsc{subs?}(A,B)$. Because of different kinds of potential relations among concepts, currently three kinds of relations are covered: subClass ($\mathcal{S}$), equivalence ($\mathcal{E}$) and disjointness ($\mathcal{D}$) axioms (see Algorithm \ref{alg3}). 

\begin{figure}[h]
\centering
\includegraphics[width=.47\textwidth]{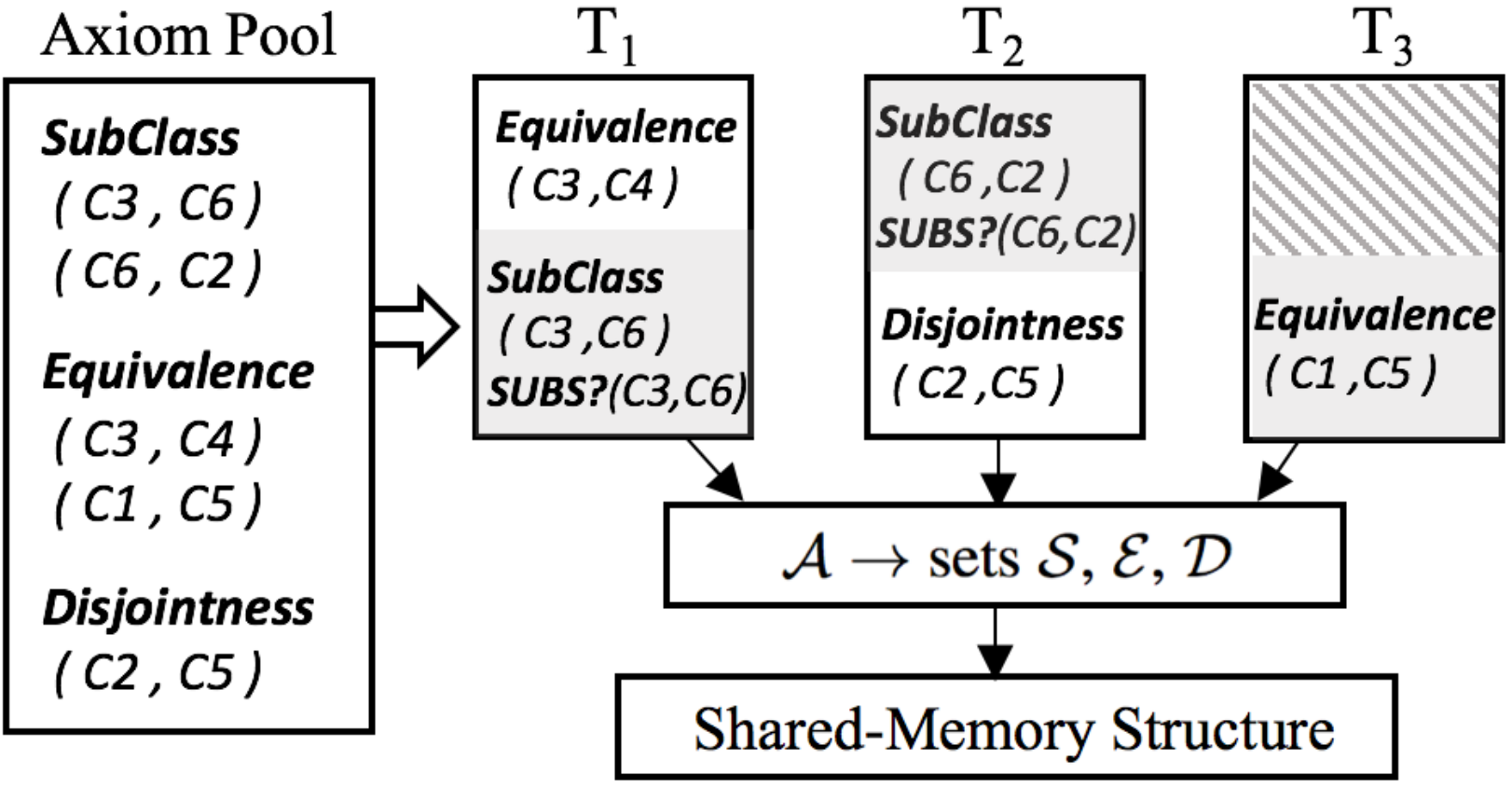}
\caption{Parallel precomputing phase}
\label{Example6}
\end{figure}

In Example \ref{ex:example-2}, the OWL input can be interpreted as shown in Figure \ref{Example6}, which has an axiom pool containing the identified axioms and three threads ($T_1, T_2, T_3$) to analyze the results. From the results shown in Figure \ref{Example2}, all the possible sets are empty, which means all the possible relations among the six satisfiable concepts have been tested or inferred and  the results are recorded in sets $\mathcal{S, E, D}$ of $\mathcal{A}$ respectively.

\SetAlgoNoLine
\begin{algorithm}[h]
       \For{each pair $\{C_i, C_j\}\in A_{i} $}{
          \If {SubClass ($C_i, C_j$)}
          {$\textsc{updateSubsumee}(C_i, C_j)$}
          \textbf{else} \If {Equivalence ($C_i, C_j$)}
          {$\textsc{updateEquivalence}(C_i, C_j)$}
          \textbf{else} \If {Disjointness ($C_i, C_j$)}
          {$\textsc{updateDisjointness}(C_i, C_j)$}
}
\caption{\textsc{preComputing}($A_i$)}
\label{alg3}
\end{algorithm}

\subsection{Classification Phase}

However, it is likely that $\mathcal{R_O}$ is not empty after the precomputing phase, then the classification phase is processed to finish the classification and guarantee the completeness of our method. Because of the wall clock time differences of subsumption tests performed by black-box reasoners, it is important to ensure concurrency and avoid longer waiting times for the remaining concepts, especially when the tests are taking longer than estimated. A work-stealing strategy is applied to schedule different threads dynamically and improve load balancing among threads to speed up the classification process.

\subsubsection{Work-Stealing Strategy} First, find all the remaining possible $\mathcal{P}_{C_i}\in \mathcal{R_O}$ with $\mathcal{P}_{C_i} \neq \emptyset$. Second, separate $\mathcal{R_O}$ into smaller subgroups $G_{i}$ and put them into a queue $Q_{i}$. The sizes of groups depend on the remaining size of $\mathcal{P}$ and the number of processors (\textit{n}) currently available. Third, if there is an idle thread available during the classification process, a new group from the queue will be given to that thread dynamically until all the subgroups have been classified and $\mathcal{R}_O$ is empty (see Algorithm \ref{alg4}). 

\SetAlgoNoLine
\SetKwRepeat{Do}{do}{while}
\SetKw{Break}{break}
\begin{algorithm}[t]
       $\textsc{enqueue}(Q_i, G_i)$\\
            \For{each pair $\{C_i, C_j\}\in Q_i$}
           {$\textsc{updateSubsumee}(C_i, C_j)$\\
           $\textsc{dequeue}(Q_i, \{C_i, C_j\})$}
        \If{$\neg\textsc{isEmpty}(Q_{i})$}
       {$\textsc{stealWork}(T, Q_i)$}
\BlankLine
\textbf{procedure} \textsc{stealWork}($T, Q_j$)\\
$\;\;\;$ \If{$\textsc{scheduleWork}(T)$}
{ 
$\;\;$ \For{each pair $\{C_m, C_n\}\in Q_j$}
{$\;\; \textsc{updateSubsumee}(C_m, C_n)$\\
$\;\; \textsc{dequeue}(Q_j, \{C_m, C_n\})$
}
}
$\;\;\;$ \If{$\neg\textsc{isEmpty}(Q_{j})$}
{$\;\; \textsc{stealWork}(T, Q_j)$}
\caption{\textsc{\textsc{classificationSubTest}$(G_i, T)$}}
\label{alg4}
\end{algorithm}

\begin{example}
Using the six concepts generated in Example \ref{ex:example-2}, all the concepts in $\mathcal{P}$ are divided into subgroups $G_i$ and put into a queue $Q$. As shown in Figure \ref{Example5}, all the generated subgroups are indicated by the colors grey or white to separate them. Suppose there are three threads ($T_1, T_2, T_3$) available, then three queues will be generated for each thread, e.g., $Q_1 = \{G_{C_{2\_1}}, G_{C_{3\_1}}\}$, $Q_2 = \{G_{C_{2\_2}}, G_{C_{3\_2}}, G_{C_{1\_2}}\}$, $Q_3 = \{G_{C_{2\_3}}, G_{C_{1\_1}}, G_{C_{4\_1}}, G_{C_{5\_1}}\}$. During the classification, when all tasks of $Q_1$ assigned in $T_1$ have finished, a task $G_{C_{5\_1}}$ (see Figure \ref{Example5}, in darker grey) still needs to be performed by $T_3$, which is currently working on $G_{C_{4\_1}}$. Therefore, the task $G_{C_{5\_1}}$ will be stolen from $T_3$ and reallocated to $T_1$. Accordingly, all the updated information will be recorded in $\mathcal{A}$ as well.

\begin{figure}[h]
\centering
\includegraphics[width=.43\textwidth]{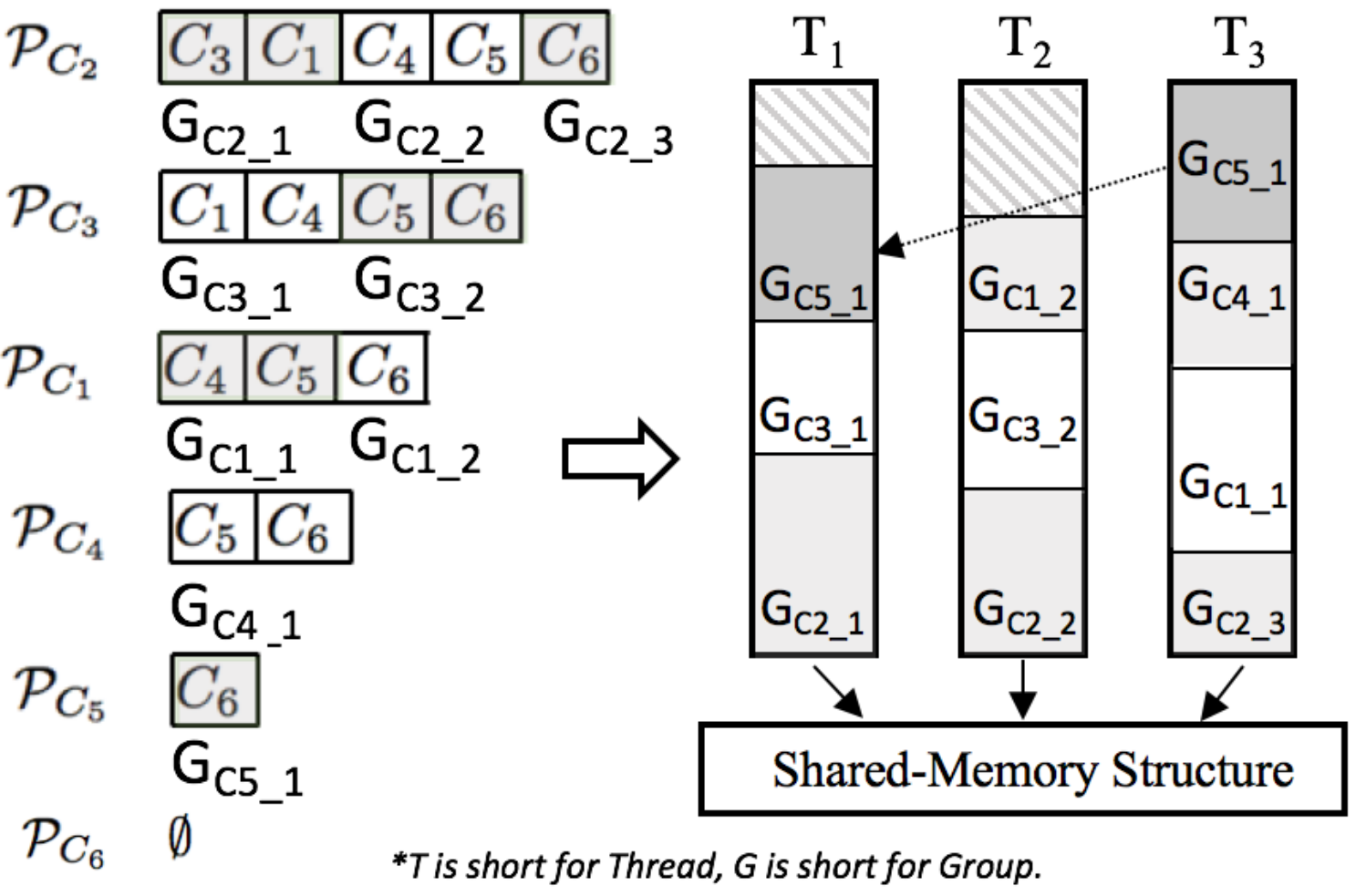}
\caption{\textit{Work-Stealing} strategy applied between $T_1$ and $T_3$}
\label{Example5}
\end{figure}

After classification, all the relevant information of each concept $\mathcal{C}_i$ is recorded in $\mathcal{A}_{C_i}$. According to $ \mathcal{A}_{C_i}$, the whole taxonomy of ontology $\mathcal{O}$ has been computed. 
\end{example}

\begin{theorem}[Soundness]
Let $\mathcal{A}_i, \mathcal{P}_{C_i}$ be a complete set for concept $C_i$ and $\mathcal{A}_j, \mathcal{P}_{C_j}$ for concept $C_j$. If the subsumption relations between a pair \{$C_i, C_j$\} are correctly inferred by sound black-box reasoners, e.g., \textsc{subs?}$(C_i, C_j)$ and \textsc{subs?}$(C_j, C_i)$, or by our algorithms \textsc{maintainingSets} (see Algorithm \ref{alg2}), which do not conclude a wrong subsumption relation between two concepts, then this parallel method is sound for $\mathcal{O}$.
\end{theorem}

\begin{theorem}[Completeness] For all the satisfiable concepts $\mathcal{C}_i \in N_\mathcal{O}$, both $\mathcal{A}_i$ and $\mathcal{P}_{C_i}$ of $\mathcal{C}_i$ are created and populated exhaustively. All the possible relations among concepts are recorded in $\mathcal{P}$. A subsumption test for each pair \{$C_i, C_j$\}  ($i \neq j$) is performed either by a complete black-box reasoner via $\textsc{subs?}(\mathcal{C}_{i},\mathcal{C}_{j})$ and $\textsc{subs?}(\mathcal{C}_{j},\mathcal{C}_{i})$ or by maintaining sets (see Algorithm \ref{alg2}). Therefore, the set $\mathcal{R_O}$ is empty if and only if all the possible relations in the sets $\mathcal{P}_{C_i}$ have been derived.
\end{theorem}

\begin{table*}[t]
\centering\
\rowcolors{2}{gray!26}{white}
\captionsetup{justification=centering}
\caption{Precomputing metrics using Hermit (wall clock time (WCT) in seconds, Equi = Equivalence Axioms,\\ Dis = Disjointness  Axioms, timeout (TO) = 1,000 seconds, \# = number of threads)}
\label{table5}
\centering
\renewcommand{\arraystretch}{1.08}
\resizebox{\textwidth}{!}{
\begin{tabular}{l|r|r|r|r|r|r|r|r|r|r}
\hline
\rowcolor{white}\multicolumn{1}{c|}{}&\multicolumn{1}{c|}{}&\multicolumn{1}{c|}{}&\multicolumn{1}{c|}{}&\multicolumn{1}{c|}{}&\multicolumn{1}{c|}{}&\multicolumn{5}{c}{\textbf{Precomputing WCT}}\\ 
\cline{7-11}
\rowcolor{white}\multicolumn{1}{c|}{\textbf{Ontology}}&\multicolumn{1}{c|}{\textbf{Axioms}}&\multicolumn{1}{c|}{$\textbf{Concepts}$}&\multicolumn{1}{c|}{$\textbf{SubClasses}$}&\multicolumn{1}{c|}{$\textbf{Equi}$}&\multicolumn{1}{c|}{$\textbf{Dis}$}&\multicolumn{1}{c|}{\textbf{\#1}}&\multicolumn{1}{c|}{\textbf{\#20}}&\multicolumn{1}{c|}{\textbf{\#60}}&\multicolumn{1}{c|}{\textbf{\#100}}&\multicolumn{1}{c}{\textbf{\#120}}\\
\hline
microbial.type & 13,584 & 4,636 & 7,255 & 935 & 31 & 304 & 93.9 & 17.5 & \textbf{0.5}  & 3.2 \\
MSC\_classes & 15,092 & 5,559 & 8,220 & 930 & 382 & 156 & 61.9  & 9.7  & \textbf{1.46}  & 2.83 \\
CURRENT& 26,374 & 6,595 & 17,180 & 2,297 & 218 & 391 & 182 & 10.9 & 8.7 & \textbf{0.74} \\
natural.product & 169,498 & 9,463 & 12,370 & 0 & 56,192 & 67.9  & 23.6  & 8.91  & 3.48  &  \textbf{2.16}  \\
vertebrate & 94,564 & 18,092 & 71,579 & 4,428  & 0 & TO  & TO & TO  & TO  &  TO \\
pr\_simple & 149,568 & 59,006 & 89,854 & 0 & 693 & TO & 464 & 147 & 105 & \textbf{38.2} \\
attributes & 221,783 & 62,035& 141,224 & 18,029 & 137 & TO & 860 & 630 & 218 & \textbf{120} \\
CLASSIFIED & 169,155 & 83,036 & 55,046 & 30,363 & 693 & TO & 883 & 570 & 236 & \textbf{79.2} \\
behavior & 354,825 & 99,360 & 241,046 & 14,013  & 62 & TO & TO & 972 & 729 & \textbf{303} \\
havioredit & 354,971 & 99,399 & 241,140 & 14,026 & 62 & TO & TO & TO & TO & TO \\
\hline
\end{tabular}
}
\end{table*}

\section{Evaluation}
Our parallel framework implements a half-matrix shared-memory structure with \textit{Atomic} from the Java Concurrent package, which supports generating more than one thread to maximize CPU utilization. The chosen black-box OWL reasoner is used  for deciding concept satisfiability and subsumption between a pair of concepts. The evaluation was performed by exclusively using an HP DL580 Scientific Linux SMP server with four 15-core processors and a total of 1 TB RAM (each processor has 256 GB of shared RAM and its 15 cores support hyperthreading). The test ontologies were selected from the ORE \shortcite{ore-2014} repository to evaluate the performance of our parallel approach. They vary by the number of axioms, concepts, and for precomputing by the number of subclass, equivalence and disjointness axioms. Considering the implementations of different reasoners and their Java compatibility, currently we successfully applied our parallel reasoning framework by using two OWL reasoners as black-box reasoners: (i) Hermit 1.3.8 \cite{reasoner-Hermit} is an OWL reasoner fully supporting OWL datatypes; and (ii) JFact 5.0.3 is a Java port of FaCT++ \cite{reasoner-FaCT++}, a tableau based OWL reasoner. For reasons of compatibility and performance, in this paper we focus on the comparison and evaluation with Hermit.

\begin{table*}
\centering
\rowcolors{2}{gray!26}{white}
\captionsetup{justification=centering}
\caption{Subsumption test metrics using Hermit (wall clock time (WCT)  in seconds, timeout (TO) = 1,000 seconds,\\ Con = Concepts, Dev = Deviation, Med = Median, Ave = Average, P(ara) = Parallel, W = without work-stealing, Her = Hermit)}
\label{table2}
\renewcommand{\arraystretch}{1.12}
\resizebox{\textwidth}{!}{
\begin{tabular}{ l|r|r|r|r|r|r|r|r|r|r|r|r}
\hline
\rowcolor{white}\multicolumn{1}{c|}{}&\multicolumn{1}{c|}{}&\multicolumn{1}{c|}{}&\multicolumn{5}{c|}{\textbf{Subsumption Test Statistics}}&\multicolumn{3}{c|}{\textbf{WCT}}&\multicolumn{2}{c}{\textbf{Speedup}}\\
\cline{4-8}\cline{9-11}\cline{12-13}
\rowcolor{white}\multicolumn{1}{c|}{\textbf{Ontology}}&\multicolumn{1}{c|}{\textbf{Axioms}}&\multicolumn{1}{c|}{\textbf{Con}}&\multicolumn{1}{c|}{\textbf{Dev}}&\multicolumn{1}{c|}{\textbf{Max}}&\multicolumn{1}{c|}{\textbf{Min}}&\multicolumn{1}{c|}{\textbf{Med}}&\multicolumn{1}{c|}{\textbf{Ave}}&\multicolumn{1}{c|}{\textbf{Para}}&\multicolumn{1}{c|}{\textbf{PW}}&\multicolumn{1}{c|}{\textbf{Her}}&\multicolumn{1}{c|}{\textbf{PW}}&\multicolumn{1}{c}{\textbf{Her}}\\
\hline
mfoem.emotion & 2,389 & 902 & 0.26 & 1.92 & 0.03 & 0.22 & 0.12& \textbf{2.7} & 25.3 & 42.1 & 9.4 & 15.6 \\
nskisimple & 4,775 & 1,737 & 0.07 & 0.42 & 0.0001 & 0.23 & 0.03 & \textbf{2.9} & 28.1  & 29.3 & 9.7 &10.1\\
geolOceanic & 6,573 & 2,324 & 0.21 & 1.01 & 0.017 & 0.55 & 0.07 & \textbf{1.4} & 19.2  &12.1 & 13.7 &8.6  \\
stateEnergy & 10,270 & 3,018 & 1.94 & 9.78 & 0.07 & 1.12 & 0.25 & \textbf{12.3} & 201  &72.9 & 16.3 &5.9 \\
aksmetrics & 11,134 & 3,889 & 0.73 & 1.24 & 0.005 & 0.34 & 0.21 & \textbf{3.3} & 46.5  &13.6 & 14.1 &4.2 \\
microbial.type & 13,584 & 4,636 & 2.15 & 13.6 & 0.03 & 3.38 & 1.13 & \textbf{26.7} & 512  &308 & 19.2 &11.5 \\
MSC\_classes & 15,092 & 5,559 & - & TO & - & - & - & TO & TO  & TO  & - & -\\
CURRENT& 26,374 & 6,595 & 10.9 & 32.8 & 0.01 & 6.34 & 2.72 & \textbf{112} & 783  & 452 & 6.9 &4.0 \\
compatibility & 21,720 & 7,929 & 4.37 & 9.38 & 0.005 & 3.72 & 0.98 & \textbf{20.2} & 240 &  22.5  & 11.9 & 1.1 \\
natural.product & 169,498 & 9,463 & 12.5 & 87.2 & 0.02 & 15.8 & 6.93 & 98.7 & 352  &\textbf{11.2} & 3.6 &0.11  \\
havioredit& 354,971 & 99,399 & - & TO & - & - & - & TO  & TO & TO  &  - & -\\
\hline
\end{tabular}
}
\end{table*}

\subsection{Benchmarks}

We tested our parallel framework against Hermit measuring the wall clock time that is separately recorded for both the precomputing and whole classification phase. The current experimental environment allows us to use up to 120 threads. The actual number of threads depends on the ontology's size and reasoning difficulty. All the experiments were repeated five times and the resulting average is used to determine wall clock times and speedup factors. Table \ref{table5} shows the characteristics of 10 selected ontologies such as the number of axioms, named concepts, subclass, equivalence and disjointness axioms. In the precomputing phase, an axiom pool is created to contain all the axioms eligible for precomputing. Axiom preprocessing is parallelized using the maximum number of threads allowed. In order to test the performance of precomputing, we tested both the sequential and parallel cases using different numbers of threads (20, 60, 100, 120). The results (wall clock time (WCT) in seconds) are shown in the five rightmost columns of Table \ref{table5}.
The best result is indicated for each ontology in bold.

In order to better assess the impact of the overhead due to parallelization and other potential factors such as the efficiency of the selected black-box reasoner, we also recorded time statistics of subsumption tests performed by the black-box reasoner, such as deviation, maximum, minimum, median, and average time. Table \ref{table2} reports various time metrics and data for 11 different ontologies over the whole classification process. The five rightmost columns present the wall clock times of our system with (Para) and without using work-stealing (PW), the times of Hermit (Her), and the speedup factors (Speedup), which are calculated by dividing the wall clock times of PW by Para and Her by Para. The best results are indicated in bold. Our results for a larger set of 40 ontologies are shown in Table \ref{tab:all-results}.

\subsection{Discussion}

Table \ref{table5} shows that the precomputing time could be significantly improved due to parallelization by using up to 120 threads (a bold font indicates the best time). The ontologies \textit{microbial.type} and \textit{CURRENT} could be processed about 600 times faster  compared to the sequential case when 100 or 120 threads are applied. The ontology \textit{vertebrate} timed out even for 120 threads due to  the black-box reasoner that is already used in the precomputing phase. The next four bigger ontologies timed out if only one thread is used but could be processed with an increasing number of threads and lead to a speedup of more than 1,000 compared to the sequential case. The biggest ontology \textit{havioredit} still timed out for 120 threads. Due to the use of parallelization and a atomic half-martix shared-memory structure together with the maximum number of available threads, a better performance is achieved by updating accumulative information and reducing the total number of subsumption tests, which results in a decreased wall clock time in the precomputing phase compared to the sequential case.

Table \ref{table2} indicates two important factors affecting the performance of our system: the partition size and the efficiency of subsumption tests. A reasonable partition size for each thread can reduce the overhead of waiting or updating information in our atomic half-matrix structure, e.g., \textit{mfoem.emotion} and \textit{nskisimple} are more than 10 times faster than Hermit when 80-100 threads are applied since each thread has a reasonable partition size and less overhead according to the deviation that is closer to the average time. 
When the size of ontologies increases, such as for \textit{geolOceanic, stateEnergy}, and \textit{aksmetrics}, a better performance is achieved with 100-120 threads because of reasonable partition sizes and uniformity of subsumption tests, which result in smaller differences between deviation and average time. 

In order to better assess the impact of black-box reasoners on our framework, we computed more statistics on subsumption tests that are also shown in Table \ref{table2}. The statistics lists 9.78s as maximum time for \textit{stateEnergy}. Thus, the performance of our framework cannot be below that maximum time. \textit{MSC\_classes} times out for Hermit and our framework. Our individual subsumption tests are performed by the black-box reasoners, and its effectiveness also constrains the performance of our framework, i.e., if a single subsumption test times out as indicated for \textit{MSC\_classes}, then our system times out also due to the black-box reasoner.
For the ontology \textit{microbial.type}, many subsumptions can be derived during parallel precomputing, which results in a speedup factor of almost 600 (see Table \ref{table5}). Moreover, if tested sequentially, this ontology requires some difficult tests which take more time than the  maximum of 13.6s (parallel testing). Due to parallel processing and fast accumulation and synchronous updating of concept relations in our atomic structure, our system can avoid these difficult tests, which makes our framework more than 10 times faster than the black-box reasoner.

When the size of ontologies increases even more, such as for \textit{CURRENT}, where many subsumptions can be derived during parallel precomputing, we achieved a speedup of 4 with 120 threads. For \textit{compatibility}, which has about 8,000 concepts, the performance of our approach is below but close to the black-box reasoner, because some subsumption tests could be avoided by black-box reasoner optimizations but were required for our framework in order to guarantee completeness. 
However, for the second last ontology \textit{natural.product} our system cannot compete with Hermit because the maximum subsumption time is very high and it seems that black-box reasoner optimizations, which are inaccessible to our system due to the black-box approach, can avoid this test. The last ontology \textit{havioredit}, which is the biggest one we chose, times out for all the reasoners and our framework. Each thread is overloaded by the number of concepts to classify, which results in more overhead in the whole classification process and, thus, causes a timeout for \textit{havioredit}, even though the precomputing phase becomes faster. 
Overall, our optimized parallel framework achieves a better performance than Hermit when enough threads are available to ensure reasonable partitions for different ontology sizes, especially if the number of concepts is less than 10,000. 

In order to solve the problem of load balancing, we used a work-stealing strategy to balance variations in partitions and subsumption tests. Therefore, Table \ref{table2} shows the wall clock times of our parallel framework without applying work stealing (PW) and in the second last column the speedup factors defined by $\mathit{\frac{PW}{Para}}$. From the results, the best performances have a factor of 19.2 and 16.3 for the ontology \textit{microbial.type} and \textit{stateEnergy} respectively, which have a high maximum time compared to the wall clock time of Para. Most of the improved speedup factors are in the range of 9-15, which show the improvements when the work-stealing strategy is applied in our approach.

\section{Conclusion}
We presented a parallel classification framework for OWL/DL reasoners, using an atomic half-matrix shared-memory structure and parallel techniques for classification. The evaluation shows that our approach is mostly more efficient for ontologies which have less than 10,000 concepts. The evaluation results indicate that if our framework would use a different and more efficient black-box reasoner, it could scale better for more difficult and/or bigger ontologies. In our future work we plan to extend our framework to be more flexible and to employ more black-box reasoners in order to improve scalability. In addition, we tested 40 more ontologies and the results are shown in Table \ref{tab:all-results}. A more detailed presentation of this approach is given in Quan \shortcite{Quan-PhD-2019} and project link \shortcite{project}.

\begin{table*}[p]
\rowcolors{2}{gray!26}{white}
\captionsetup{justification=centering}
\caption{Time Metrics of tested OWL ontologies using parallel framework\\ (wall clock time (WCT) in seconds, timeout (TO) = 1000 seconds, Sequen = Sequential, Para = Parallel Framework)}
\label{tab:all-results}
\centering
\renewcommand{\arraystretch}{1.19}
\resizebox{\textwidth}{!}{
\begin{tabular}{ r|l|r|r|r|r|r|r|r}
\hline
\rowcolor{white}\multicolumn{1}{c|}{}&\multicolumn{1}{c|}{}&\multicolumn{1}{c|}{}&\multicolumn{1}{c|}{}&\multicolumn{2}{c|}{\textbf{Precomputing}}&\multicolumn{2}{c|}{\bf{WCT}}&\multicolumn{1}{c}{\textbf{Speedup}}\\ 
\cline{5-6}\cline{7-8}
\rowcolor{white}\multicolumn{1}{c|}{\textbf{\#}}&\multicolumn{1}{c|}{\textbf{Ontology}}&\multicolumn{1}{c|}{\textbf{Concept}}&\multicolumn{1}{c|}{\textbf{Expressivity}}&\multicolumn{1}{c|}{\textbf{Sequen}}&\multicolumn{1}{c|}{\textbf{Parallel}}&\multicolumn{1}{c|}{\textbf{Para}}&\multicolumn{1}{c|}{\textbf{Hermit}}&\multicolumn{1}{c}{\textbf{Factor}}\\
\hline
1& SocialUnits & 156 &$\mathcal{SHOIN(D)}$& 84.34 & 0.86 & 16.09 & 353.52 & 21.97 \\
2& 00021& 156 &$\mathcal{SHOIN(D)}$& 105.98 & 0.91 & 15.43 & 260.42 & 16.88  \\
3& rnao.owl & 240 &$\mathcal{SRIQ}$ & 1.16 & 0.29 & 3.06 & 109.99 & 35.94 \\
4& tionmodule & 256 & $\mathcal{SHOIN(D)}$& 523.7 & 1.12 & 640 & 909.5 & 1.42 \\
5& genetic& 386 & $\mathcal{SROIQ(D)}$& 671.9 & 8.80 & 31.17 & 530.31 & 17.01 \\
6& WM30 & 415 & $\mathcal{SROIQ(D)}$& TO & 0.74 & TO & 798.33 & - \\
7& ainability & 824 & $\mathcal{SHOIN(D)}$& 6.06 & 0.14 & 0.91 & 15.41 & 16.93  \\
8& sadiobjects & 828 & $\mathcal{ALN(D)}$& 0.66 & 0.49 & 2.53 & 4.42 & 1.75 \\
9& Microbiota & 868 & $\mathcal{SHOIN(D)}$& 6.36 & 0.19 & 0.97 & 17.94 & 18.49 \\
10& mfoem.emotion & 902 & $\mathcal{SROIQ}$& 35.11 & 0.88 & 2.77 & 42.11 & 15.20  \\
11& onsumption & 945&$\mathcal{ALCHOIQ(D)}$ & 233.71 & 1.21 & 2.19 & 20.82 & 9.51  \\
12& emistrycomplex & 1,041& $\mathcal{SHIQ(D)}$ & 12.43 & 0.61 & 8.53 & 14.19 & 1.66 \\
13 & nskisimple & 1,737& $\mathcal{SRIQ(D)}$ & 36.32 & 0.21 & 2.9 & 29.3 & 10.10  \\
14 & Earthquake & 2,013 & $\mathcal{SHOIN(D)}$ & 20.53 & 0.68 & 7.73 & 14.77 & 1.91 \\
15 & geolOceanic & 2,324 & $\mathcal{SHOIN(D)}$& 23.80 & 0.55 & 1.38 & 12.03 & 8.72  \\
16 & landCoastal & 2,660 & $\mathcal{SHOIN(D)}$ & 29.02 & 0.70 & 1.48 & 17.22 & 11.64 \\
17 & mergedobi & 2,638 & $\mathcal{SHOIN(D)}$& 351.05 & 0.96 & TO & 364.58 & -  \\
18 & 00350 & 2,638 & $\mathcal{SHOIN(D)}$& 441.31 & 3.25 & 28.17 & 310.32 & 11.02  \\
19 & obi & 2,750 & $\mathcal{SROIQ(D)}$& 336.96 & 2.49 & 35.3 & 342.98 & 9.72  \\
20 & quanSpace & 2,999 & $\mathcal{SHOIN(D)}$& 145.83 & 0.42 & 38.21 & 380.11 & 9.95  \\
21 & EnergyFlux & 3,008 & $\mathcal{SHOIN(D)}$& 193.26 & 0.36 & 121 & 277.36 & 2.29  \\
22 & stateEnergy & 3,018 & $\mathcal{SHOIN(D)}$& 131.84 & 0.99 & 12.26 & 72.92 & 5.95  \\
23 & rDataModel & 3,049 & $\mathcal{SHOIN(D)}$& 136.57 & 1.32 & 68.3 &757.11 & 11.09  \\
24 & virControl & 3,274 & $\mathcal{SHOIN(D)}$ & 164.98 & 0.42 & 45.6 & 439.83 & 9.65 \\
25 & aksmetrics & 3,889 & $\mathcal{SHIQ(D)}$& 6.43 & 0.6 & 3.25 & 13.64 & 4.20  \\
26 & microbial.type & 4,636 & $\mathcal{SROIQ(D)}$& 304.34 & 0.51 & 26.67 & 308.73 & 11.58  \\
27 & MSC\_classes & 5,559 & $\mathcal{ALCQ}$ & 156.84 & 1.46 & TO & TO & - \\
28 & obo.PREVIOUS & 6,580 & $\mathcal{SRIQ}$& 378.09 & 0.55 & 311.8 & 646.11 & 2.07  \\
29& obo.CURRENT & 6,595 & $\mathcal{SRIQ}$& 391.48 & 0.74 & 112.5 & 452.5 & 4.02 \\
30 & PREVIOUS & 7,335 & $\mathcal{SRIQ}$& TO & 1.79 & TO & TO & -  \\
31& SMOtop & 7,782 & $\mathcal{SHOIN(D)}$& TO & 32.41 & 432.7 & TO & 2.31 \\
32& COSMO & 7,804 & $\mathcal{SHOIN(D)}$ & TO & 33.42 & 728.6 & TO & 1.37 \\
33& compatibility & 7,929 & $\mathcal{ALCIQ(D)}$& 37.78 & 0.63 & 20.15 & 22.23 & 1.10 \\
34& EnzyO & 8,223 & $\mathcal{ALUIN(D)}$& TO & 1.74 & TO &TO & - \\
35& natural.product & 9,463 & $\mathcal{SHOIN(D)}$& 67.89 & 2.16 & 98.72 &11.21 & 0.11  \\
36& vertebrate & 18,092 & $\mathcal{SRIQ}$ & TO & TO & TO & TO & -\\
37& temetazoan & 32,750& $\mathcal{SRIQ}$ & TO & TO & TO & TO & -  \\
38& ewasserted & 63,848 & $\mathcal{SRIQ}$& TO & 53.9 & TO & TO & -  \\
39& ersections & 70,232 & $\mathcal{SRIQ}$& TO & TO & TO & TO & - \\
40& havioredit & 99,399 & $\mathcal{SRIQ}$& TO & TO & TO & TO & -  \\
\hline
\end{tabular}
}
\end{table*}

\clearpage
\bibliographystyle{named}
\bibliography{reference}

\end{document}